\documentclass[journal,final,letterpaper,onecolumn]{IEEEtran}

\usepackage[textfont=normalfont,footnotesize]{caption}
\newcommand{\mytblsz}{\small}
\newcommand{\mytblszbegin}{\begin{small}}
\newcommand{\mytblszend}{\end{small}}

\usepackage{epstopdf}
\usepackage{fullpage,times,color}
\usepackage[ruled]{algorithm2e}
\usepackage{graphicx}
\usepackage{epsfig}
\usepackage{hyperref}
\usepackage{multirow}

\usepackage{tikz}
\usetikzlibrary{calc}

\graphicspath{{./}} 

\usepackage{amsmath,amssymb}

\newtheorem{assumption}{Assumption}

\newcommand{\mv}[1]{{\mathbf #1}}  %% matrix and vector 
\newcommand{\Indnoarg}{{\mathcal I}}

\newcommand{\z}{\hspace{0.2in}}

\newcommand{\totalloss}{{\cal Q}}

\newcommand{\func}{h}

\newcommand{\totalreg}{{\cal R}}  % regularization term 

\newcommand{\funcforest}{\func_{\forest}}
\newcommand{\functree}{\func_{\tree}}
\newcommand{\reg}{r}

\newcommand{\bx}{{\mathbf x}}
\newcommand{\cH}{{\mathcal H}}
\newcommand{\cR}{{\mathcal R}}
\newcommand{\cL}{{\mathcal L}}
\newcommand{\Real}{{\mathbb R}}
\newcommand{\cC}{{\mathcal C}}
\newcommand{\cA}{{\mathcal A}}

\newcommand{\forest}{{\mathcal F}}  
\newcommand{\thv}{t}   % threshold value 
\newcommand{\basis}{{b}} % basis function 
\newcommand{\weight}{\alpha}

\newcommand{\assign}{{\leftarrow}}

\newcommand{\rgfLeaf}{RGF-$L_2$}
 
\newcommand{\regMiner}{min-penalty regularizer} 
\newcommand{\regMiners}{min-penalty regularizers} 

\title{Learning Nonlinear Functions Using Regularized Greedy Forest\thanks{
To appear in IEEE Transactions on Pattern Analysis and Machine Intelligence, 36(5):942--954, May 2014. 
\copyright 2014 IEEE. Personal use of this material is permitted. Permission from IEEE must be obtained for all other uses, in any current or future media, including reprinting/republishing this material for advertising or promotional purposes, creating new collective works, for resale or redistribution to servers or lists, or reuse of any copyrighted component of this work in other works.
}}
\author{
Rie Johnson \\
RJ Research Consulting \\
\and 
Tong Zhang \\
Rutgers University}

\begin{document}
%\sloppy
\SetEndCharOfAlgoLine{}

\maketitle
\begin{abstract}
We consider the problem of learning a forest 
of nonlinear decision rules with general loss functions.  
The standard methods employ boosted decision trees such as Adaboost for
exponential loss and Friedman's gradient boosting for general loss.
In contrast to these
traditional boosting algorithms that treat a tree learner as a black box, the method we propose directly learns decision forests via fully-corrective regularized greedy search using the underlying forest structure.  
Our method achieves higher accuracy and smaller models than gradient boosting 
on many of the datasets we have tested on.  
\end{abstract}


\section{Introduction}

Many application problems in machine learning require learning nonlinear functions from data. 
A popular method to solve this problem is through decision tree learning
(such as CART \cite{BrFrOlSt84} and C4.5 \cite{Quinlan93}), which has an important advantage for handling heterogeneous data with ease when different features come from different sources. This makes decision trees a popular ``off-the-shelf'' machine learning method that can be readily applied to any data without much tuning; in comparison, alternative algorithms such as neural networks require significantly more tuning. However, a disadvantage of decision tree learning is that it does not generally achieve the most accurate prediction performance, when compared to other methods. 
A remedy for this problem is through {\em boosting} \cite{FS97,Friedman:01a,Schapire:03a}, where one builds an additive model of decision trees by sequentially building trees one by one. In general ``{\em boosted decision trees}'' is regarded as the most effective off-the-shelf nonlinear learning method
for a wide range of application problems. 

In the boosted tree approach, one considers an additive model over multiple decision trees, and thus, we will refer to the resulting function as a {\em decision forest}. Other approach to learning decision forests include {\em bagging} and {\em random forests} \cite{Breiman:1996,Breiman:01a}.
In this context, we may view boosted decision tree algorithms as methods to learn decision forests by 
applying a greedy algorithm (boosting) on top of a decision tree base learner. This indirect approach is sometimes referred to as 
a {\em wrapper} approach (in this case, wrapping boosting procedure over decision tree base learner); the boosting wrapper simply treats the decision tree base learner as a black box and it does not take advantage of the tree structure itself. 
The advantage of such a wrapper approach is that the underlying base learner can be changed to other procedures with
the same wrapper; the disadvantage is that for any specific base learner which may have additional structure to explore, a generic wrapper
might not be the optimal aggregator.

Due to the practical importance of boosted decision trees in applications, it is natural to ask whether one can design a more direct procedure that specifically learns decision forests without using a black-box decision tree learner under the wrapper. The purpose of doing so is that by directly taking advantage of the underlying tree structure, we shall be able to design a more effective algorithm for learning the final nonlinear decision forest.
This paper attempts to address this issue, where we propose a direct decision forest learning algorithm called {\em Regularized Greedy Forest} or RGF. 
We are specifically interested in an approach that can handle general loss functions (while, for example, Adaboost is specific to a certain loss function), which leads to a wider range of applicability.  
An existing method with this property is {\em gradient boosting decision tree} (GBDT) \cite{Friedman:01a}.  
We show that RGF can deliver better results than GBDT on a number of datasets we have tested on.  

\section{Problem Setup}

We consider the problem of learning a single nonlinear function $\func(\bx)$ on some input vector 
$\bx=[\bx[1],\ldots,\bx[d]] \in \Real^d$ from a set of training examples.
In supervised learning, we are given a set of input
vectors $X=[\bx_1,\ldots,\bx_n]$ with labels $Y=[y_1,\ldots,y_m]$
(here $m$ may not equal to $n$).
Our training goal is to find a nonlinear prediction function 
$\hat{\func}(\bx)$ 
from a function class $\cH$
that minimizes a risk function
\begin{equation}
  \hat{\func}=\arg\min_{\func \in \cH}  \cL(\func(X),Y)  .
  \label{eq:genloss}
\end{equation}
Here $\cH$ is a pre-defined nonlinear function class,
$\func(X)=[\func(\bx_1),\ldots,\func(\bx_n)]$ is a vector of size $n$, and 
$\cL(\func, \cdot)$ is a general loss  function of vector $h \in \Real^n$.

The loss function $\cL(\cdot,\cdot)$ is given by the underlying problem. For example, for regression problems,
we have $y_i \in \Real$ and $m=n$. If we are interested in the conditional mean of $y$ given $\bx$, 
then the underlying loss function corresponds to least squares regression as follows:
\[
\cL(\func(X),Y)= \sum_{i=1}^n (\func(\bx_i) - y_i)^2 .
\]
In binary classification, we assume that $y_i \in \{\pm 1\}$ and $m=n$. We may consider the
logistic regression loss function as follows:
\[
  \cL(\func(X),Y)= \sum_{i=1}^n \ln (1 + e^{-\func(\bx_i) y_i}) .
\]
Another important problem that has drawn much attention in recent years is the pair-wise preference learning
(for example, see \cite{HerGraObe00,FISS03}),
where the goal is to learn a nonlinear function $\func(\bx)$ so that
$\func(\bx)>\func(\bx')$ when $\bx$ is preferred over $\bx'$.
In this case, $m=n(n-1)$, and the labels encode pair-wise preference
as $y_{(i,i')}=1$ when $\bx_i$ is preferred over $\bx_{i'}$, and 
$y_{(i,i')}=0$ otherwise. 
For this problem, we may consider the following loss function that suffers a loss 
when $\func(\bx) \leq \func(\bx')+1$. That is, the formulation encourages the separation of 
$\func(\bx)$ and $\func(\bx')$ by a margin when $\bx$ is preferred over $\bx'$:
\begin{align*} 
  \cL(\func(X),Y) =& \sum_{(i,i'): y_{(i,i')}=1} \max(0, 1-(\func(\bx_{i}) - \func(\bx_{i'})))^2 . 
\end{align*}

\begin{figure}[htb]
  \centering
  \begin{tikzpicture}
  [
    scale=1.2,
    observed/.style={circle,inner sep=2.5pt,draw=black,fill=black!10},
    front/.style={circle,inner sep=2.5pt,draw=black,fill=black!0}
   ]

   \node[observed,name=T1] at (1,0){\tiny root};
   \node [observed,name=T11] at (2,0.5) {};
   \draw[->] (T1) to (T11);
   \node [observed,name=T12] at (2,-0.5) {};
   \draw[->] (T1) to (T12);
  \node [observed,name=T121] at (3,0) {};
   \draw[->] (T12) to (T121);
   \node [observed,name=T122] at (3,-1) {};
   \draw[->] (T12) to (T122);
   \node [observed,name=T1211] at (4,0.5) {};
   \draw[->] (T121) to (T1211);
   \node [observed,name=T1212] at (4,-0.5) {};
   \draw[->] (T121) to (T1212);

\end{tikzpicture}
%%\vspace{-10pt}
  \caption{Decision Tree}
 \label{fig:rules}
\end{figure}
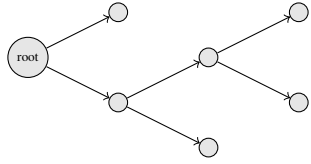

Given data $(X,Y)$ and a general loss function $\cL(\cdot,\cdot)$ in (\ref{eq:genloss}), there are two basic questions to
address for nonlinear learning. The first is the form of nonlinear function class $\cH$, and the second is
the learning/optimization algorithm. 
This paper achieves nonlinearity by using additive models of the form:
\begin{equation}
\cH = \left\{\func(\cdot): \func(\bx)= \sum_{j=1}^K \weight_j \basis_j(\bx) ; \; \forall j, \basis_j \in \cC \right\} ,
\label{eq:additive-model}
\end{equation}
where each $\weight_j \in \Real$ is a coefficient that can be optimized, and
each $\basis_j(\bx)$ is by itself a nonlinear function (which we may refer to as a nonlinear basis function or an atom) 
taken from a base function class $\cC$.
The base function class typically has a simple form that can be used in the underlying algorithm.
This work considers decision rules as the underlying base function class that is of the form
\begin{equation}
\cC= \left\{ \basis(\cdot): \basis(\bx)=
\prod_{j}\Indnoarg(\bx[i_j] \leq \thv_{j}) 
                 \prod_{k}\Indnoarg(\bx[i_k] > \thv_{k})
\right\} , 
\label{eq:decision-rule}
\end{equation}
where $\{(i_j,\thv_{j}), (i_k,\thv_{k})\}$ are a set of (feature-index, threshold) pair,
and $\Indnoarg(x)$ denotes the indicator function: 
$\Indnoarg(p)=1$ if $p$ is true; $0$ otherwise.  
%%---
Decision rules can be graphically represented with a tree structure.  In Fig. \ref{fig:rules}, 
each tree edge $e$ is associated with a variable $k_e$ and threshold $\thv_e$, and 
denotes a decision of the form $\Indnoarg(\bx[k_e] \leq \thv_e)$ or $\Indnoarg(\bx[k_e] > \thv_e)$.
Each node denotes a nonlinear decision rule of the form (\ref{eq:decision-rule}), which is the product 
of decisions along the edges leading from the root to this node.

%%---

Since the space of decision rules is rather large, for computational purposes,
we have to employ a structured search over the set of decision rules.
The 
%underlying 
optimization procedure 
we propose 
is a structured greedy search algorithm which we call regularized
greedy forest (RGF). 
To introduce RGF, we first discuss pros and cons of the existing method for general loss, gradient boosting \cite{Friedman:01a}, in the next section.  

\section{Gradient Boosted Decision Tree}

Gradient boosting is a method to minimize (\ref{eq:genloss}) with additive model (\ref{eq:additive-model}) by assuming
that there exists a nonlinear base learner (or oracle) $\cA$ that satisfies Assumption~\ref{assump:base-learner}.
\begin{assumption}\label{assump:base-learner}
  A base learner for a nonlinear function class $\cA$ is a regression optimization method 
  that takes as input any pair $\tilde{X}=[\tilde{\bx}_1,\ldots,\tilde{\bx}_n]$ and
  $\tilde{Y}=[\tilde{y}_1,\ldots,\tilde{y}_n]$ and outputs a nonlinear function $\hat{g}=\cA(\tilde{X},\tilde{Y})$ that approximately solves
  the regression problem:
  \[
  \hat{\basis} \approx \arg\min_{\basis \in \cC} \min_{\beta \in \Real} \sum_{j=1}^n (\beta \cdot \basis(\tilde{\bx}_j)- \tilde{y}_j)^2 .
  \]
\end{assumption}

The gradient boosting method is a wrapper (boosting) algorithm that solves (\ref{eq:genloss}) with
a base learner $\cA$ defined above and additive model defined in (\ref{eq:additive-model}). 
%%The general algorithm is described in Algorithm~\ref{alg:gb}.
Of special interest for this paper and for general applications is the decision tree base learner, for which
$\cC$ is the class of $J$-leaf decision trees, with each node associated with a decision rule of the form (\ref{eq:decision-rule}).
In order to take advantage of the fact that each element in $\cC$ contains $J$ (rather than one) decision rules,
%%Algorithm~\ref{alg:gb} 
the gradient boosting method 
can be modified by adding a partially corrective update step that optimizes all $J$ coefficients associated with the $J$ decision rules returned by $\cA$.
This adaption was suggested by Friedman.  %% and implemented in our experimental comparisons.
We shall refer to this modification as gradient boosted decision tree (GBDT), and the details are listed in
Algorithm~\ref{alg:gbdt}.

%%--------------
\begin{algorithm}
%%$\func_0(\bx) \assign  0$ \\
$\func_0(\bx) \assign  \arg\min_\rho \cL(\rho,Y)$ \\
\For{$k=1$ \KwTo $K$}{
  $\tilde{Y}_k \assign -\partial \cL(h, Y) /\partial h|_{h=\func_{k-1}(X)}$ \\
  Build a $J$-leaf decision tree $T_k \assign \cA(X,\tilde{Y}_k)$ with leaf-nodes $\{\basis_{k,j}\}_{j=1}^J$ \\
  \lFor{$j=1$ \KwTo $J$}{
    $\beta_{k,j} \assign 
    \arg\min_{\beta \in \Real} \cL(\func_{k-1}(X)+\beta\cdot\basis_{k,j}(X),Y)$
  } 
  $\func_k(\bx) \assign \func_{k-1}(\bx)+s \sum_{j=1}^J \beta_{k,j} \cdot \basis_{k,j}(\bx)$    \z // $s$ is a shrinkage parameter \\
}
\Return $\func(\bx)=\func_K(\bx)$ \\
\caption{
Gradient Boosted Decision Tree (GBDT) \cite{Friedman:01a}
}
\label{alg:gbdt}
\end{algorithm}

Gradient boosting may be regarded as a functional generalization of gradient descent method
$h_{k} \leftarrow h_{k-1} - s_k \frac{\partial \cL(h)}{\partial h}|_{h=h_{k-1}}$, 
where the shrinkage parameter $s$ corresponds to the step size $s_k$ in gradient descent, 
and 
$-\frac{\partial \cL(h)}{\partial h}|_{h=h_{k-1}}$
is approximated using the regression tree output. 
The shrinkage parameter $s>0$ is a tuning parameter that can affect performance, as noticed by Friedman.
In fact, the convergence of the algorithm generally requires choosing $s \beta_k \to 0$ as indicated in the theoretical analysis of \cite{ZhYu05},
which is also natural when we consider that it is analogous to step size in gradient descent. This is consistent with 
Friedman's own observation, who argued 
that in order to achieve good prediction performance (rather than computational efficiency), one should take as small a step size as possible (preferably infinitesimal step size each time), and the resulting procedure is often referred to as $\epsilon$-boosting.

GBDT constructs a decision forest which is an additive model of $K$ decision trees.
The method has been very successful for many application problems, and its main advantage is that the method can automatically find nonlinear interactions via decision tree learning (which can easily deal with heterogeneous data), and it has relatively few tuning parameters for a nonlinear learning scheme (the main tuning parameters are the shrinkage parameter $s$, number of terminals per tree $J$, and 
the number of trees $K$). 
However, it has a number of disadvantages as well.  %%, which we address in this work. 
First, there is no explicit regularization in the algorithm, and in fact, it is argued in \cite{ZhYu05}
that the shrinkage parameter $s$ plus early stopping (that is $K$) interact together as a form of regularization. 
In addition, the number of nodes $J$ can also be regarded as a form of regularization. The interaction of these parameters in
terms of regularization is unclear, and the resulting implicit
regularization may not be effective.
The second issue is also a consequence of using small step size $s$ as implicit regularization.  
Use of small $s$ 
%%can lead to a huge number of trees, 
could lead to a huge model, 
which is very undesirable as it leads to high computational cost of applications (i.e., making predictions).  
Third, the regression tree learner is treated as a black box, and its only purpose is to return 
$J$ nonlinear terminal decision rule basis functions. This again may 
%be inefficient 
not be effective
because the procedure separates tree learning
and forest learning, and hence the algorithm itself is not necessarily the most 
%efficient 
effective
method to construct the decision forest.

\section{Fully-Corrective Greedy Update and Structured Sparsity Regularization}
\label{sec:fc_sparse}

As mentioned above, 
one disadvantage of gradient boosting is that according to Friedman, 
in order to achieve good performance in practice, the shrinkage parameter $s$ 
may need to be small, and he also argued for infinitesimal step size.  
This practical observation is supported by the theoretical analysis in
\cite{ZhYu05} which showed that if we vary the shrinkage $s$ for each iteration $k$ as $s_k$, then
for general loss functions with appropriate regularity conditions, the procedure converges as $k \to \infty$ if 
we choose the sequence $s_k$ such that $\sum_k s_k|\beta_k| = \infty$ and $\sum_k s_k^2 \beta_k^2 < \infty$.
This condition is analogous to a related condition for the step size
of gradient descent method which also requires the step-size to approach zero.
Fully Corrective Greedy Algorithm is a modification of Gradient Boosting that can avoid the potential small step size problem. 
The procedure is described in Algorithm~\ref{alg:fully-corrective}. 

\begin{algorithm}[H]
$\func_0(\bx) \assign  \arg\min_\rho \cL(\rho,Y)$ \\
\For{$k=1$ \KwTo $K$}{
  $\tilde{Y}_k \assign -\partial \cL(h, Y) /\partial h|_{h=\func_{k-1}(X)}$ \\
  $\basis_k \assign \cA(X,\tilde{Y}_k)$ \\
  let $\cH_k=\{ \sum_{j=1}^k \beta_j \basis_j(\bx) : \beta_j \in \Real\}$ \\
 $\func_k(\bx) \assign \arg\min_{\func \in \cH_k} \cL(\func(X),Y)$   \z // fully-corrective step \\
}
\Return $\func(\bx)=\func_K(\bx)$ \\
\caption{
  Fully-Corrective Gradient Boosting \cite{ShSrZh09}
}
\label{alg:fully-corrective}
\end{algorithm}

In gradient boosting or its variation with tree base learner of Algorithm~\ref{alg:gbdt}, 
the algorithm only does a partial corrective step that optimizes either the coefficient of the last basis function
$\basis_k$ (or the last $J$ coefficients). 
The main difference of the fully-corrective gradient boosting is the fully-corrective-step
that optimizes all coefficients $\{\beta_j\}_{j=1}^k$  for basis functions $\{\basis_j\}_{j=1}^k$ obtained so far at each iteration $k$.
It was noticed empirically that such fully-corrective step can significantly accelerate the convergence of boosting procedures
\cite{WarmuthLiRa06}. This observation was theoretically justified in \cite{ShSrZh09} where the following rate of convergence
was obtained under suitable conditions: there exists a constant $C_0$ such that
\[
\cL(\func_k(X),Y) \leq \inf_{\func \in \cH} \left[\cL(\func(X),Y) + \frac{C_0 \|\func\|_\cC^2}{k} \right] ,
\]
where $C_0$ is a constant that depends on properties of $\cL(\cdot,\cdot)$ and the function class $\cH$, and 
\[
\|\func\|_\cC = \inf \left\{ \sum_j |\alpha_j| : \func(X)=\sum_j \alpha_j \basis_j(X); \basis_j \in \cC\right\} .
\]
In comparison, with only partial corrective optimization as in the original gradient boosting, no such convergence
rate is possible. Therefore the fully-corrective step is not only intuitively sensible, but also important theoretically.
The use of fully-corrective update (combined with regularization)
automatically removes the need for using the undesirable small step $s$ needed in the 
traditional gradient boosting approach.

However, such an aggressive greedy procedure will lead to quick overfitting of the data if not appropriately regularized
(in gradient boosting, an implicit regularization effect is achieved by small step size $s$, as argued in \cite{ZhYu05}).
Therefore we are forced to impose an explicit regularization to prevent overfitting.

This leads to the second idea in our approach, which is to impose explicit regularization via the concept of {\em structured sparsity} 
that has drawn much attention in recent years 
\cite{Baraniuk08model,JeAuBa09,Jacob09icml,Bach08-hie,Bach09-nonlinear,HuangZhang09:structured_sparsity}. 
The general idea of structured sparsity is that in a situation where a sparse solution is assumed, one can take advantage of the sparsity structure underlying the task. 
In our setting, we seek a {\em sparse} combination of decision rules
(i.e., a compact model), and we have 
the forest structure to explore, which can be viewed as graph sparsity structures.  
Moreover, 
the problem can be considered as a variable selection problem.  
Search over all nonlinear
interactions (atoms) over $\cC$ is computationally difficult or infeasible; one has to impose structured search over atoms. 
The idea of structured sparsity 
is that by exploring the fact that not all sparsity patterns are equally likely, one can select appropriate variables
(corresponding to decision rules in our setting) more effectively by preferring certain sparsity patterns more than others. 
For our purpose, one may impose structured regularization and search to prefer one sparsity pattern over another, exploring the underlying forest structure.  

This work considers the special but important case of learning a forest of nonlinear decision rules; although this may be considered
as a special case of the general structured sparsity learning with an underlying graph, the problem itself is rich and important
enough and hence requires a dedicated investigation. 
Specifically, we integrate this framework with specific tree-structured regularization and structured greedy search to obtain
an effective algorithm that can 
outperform 
the popular and important gradient boosting method.
In the context of nonlinear learning with graph structured sparsity,
we note that a variant of boosting was proposed in \cite{FreMas99}, where the idea is to split trees not only at the leaf nodes, but
also at the internal nodes at every step. However, the method is prone to overfitting due to the lack of regularization, and
is computationally expensive due to the multiple splitting of internal nodes. We shall avoid such a strategy in this work.

\newcommand{\newforest}{\tilde{\forest}}
\newcommand{\tree}{T}
\newcommand{\newtree}{\tilde{T}}
\newcommand{\leafset}{L}
\newcommand{\tleafset}{\leafset_\tree}
\newcommand{\newtleafset}{\leafset_{\newtree}}
\newcommand{\ancset}[1]{A(#1)} 
\newcommand{\childset}[1]{C(#1)} 
\newcommand{\parent}[1]{p(#1)}
\newcommand{\isChildOf}[2]{\parent{#1}=#2}
\newcommand{\neighset}[1]{N(#1)}

\newcommand{\lw}{\weight}
\newcommand{\ww}{\beta}
\newcommand{\anyww}{\beta}
\newcommand{\mww}{\bar{\ww}}
\newcommand{\lam}{\lambda}
\newcommand{\bdr}{\gamma} % base for depth regularization
\newcommand{\rt}{o_{\tree}}
\newcommand{\regww}{\beta}

\newcommand{\vx}{v}
\newcommand{\wx}{w}
\newcommand{\xx}{x}

\newcommand{\lamv}[1]{\bdr^{d_{#1}}}
\newcommand{\neighlam}{ 1 + 2\bdr } 

%**************
\section{Regularized Greedy Forest}
\label{sec:alg}

The method we propose addresses the issues of the standard method GBDT described above by directly learning a decision forest via fully-corrective regularized greedy search.    
The key ideas discussed in Section \ref{sec:fc_sparse} can be summarized as follows.  

%----
First, we introduce an explicit regularization functional on the nonlinear function $\func$ and optimize
\begin{equation}
  \hat{\func}=\arg\min_{\func \in \cH}  \left[\cL(\func(X),Y)  + \cR(\func) \right] 
  \label{eq:reg-loss}
\end{equation}
instead of (\ref{eq:genloss}). 
In particular, we define regularizers that explicitly take advantage of individual tree structures.  

%-----
Second, we employ {\em fully-corrective greedy} algorithm which repeatedly re-optimizes the coefficients of {\em all} the decision rules obtained so far while rules are added into the forest by greedy search.  
Although such an aggressive greedy procedure could lead to quick overfitting if not appropriately regularized, our formulation includes explicit regularization to avoid overfitting and the problem of huge models caused by small $s$.  

%----
Third, we perform structured greedy search {\em directly} over forest nodes based on the forest structure (graph sparsity structure) employing the concept of structured sparsity.  
At the conceptual level, our nonlinear function $\func(\bx)$ is explicitly defined as an additive model on forest nodes (rather than trees) consistent with the underlying forest structure. 
In this framework, it is also possible to build a forest by growing multiple trees simultaneously.  

%-----
Before going into more detail, 
we shall introduce some definitions and notation 
that allow us to formally define the underlying formulations and procedures. 

\subsection{Definitions and notation}

\begin{figure}[htb]
  \centering
  \begin{tikzpicture}
  [
    scale=1.2,
    observed/.style={circle,inner sep=2.5pt,draw=black,fill=black!10},
    front/.style={circle,inner sep=2.5pt,draw=black,fill=black!0}
   ]

   \node [observed,name=root] at (0,0) {\tiny root};

   \node[observed,name=T1] at (1,0){\tiny $T_1$};
   \draw[->,dashed] (root) to (T1);
   \node [observed,name=T11] at (2,0.5) {};
   \draw[->] (T1) to (T11);
   \node [observed,name=T12] at (2,-0.5) {};
   \draw[->] (T1) to (T12);
  \node [observed,name=T121] at (3,0) {};
   \draw[->] (T12) to (T121);
   \node [observed,name=T122] at (3,-1) {};
   \draw[->] (T12) to (T122);
   \node [observed,name=T1211] at (4,0.5) {};
   \draw[->] (T121) to (T1211);
   \node [observed,name=T1212] at (4,-0.5) {};
   \draw[->] (T121) to (T1212);

  \node[observed,name=T2] at (0,1){\tiny $T_2$};
   \draw[->,dashed] (root) to (T2);
   \node [observed,name=T21] at (0.5,2) {};
   \draw[->] (T2) to (T21);
   \node [observed,name=T22] at (-0.5,2) {};
   \draw[->] (T2) to (T22);
   \node [observed,name=T221] at (0,3) {};
   \draw[->] (T22) to (T221);
   \node [observed,name=T222] at (-1,3) {};
   \draw[->] (T22) to (T222);

   \node[observed,name=T3] at (-1,0){\tiny $T_3$};
   \draw[->,dashed] (root) to (T3);
   \node [observed,name=T31] at (-2,0.5) {};
   \draw[->] (T3) to (T31);
   \node [observed,name=T32] at (-2,-0.5) {};
   \draw[->] (T3) to (T32);

\end{tikzpicture}
%%\vspace{-10pt}
  \caption{Decision Forest}
 \label{fig:forest}
\end{figure}
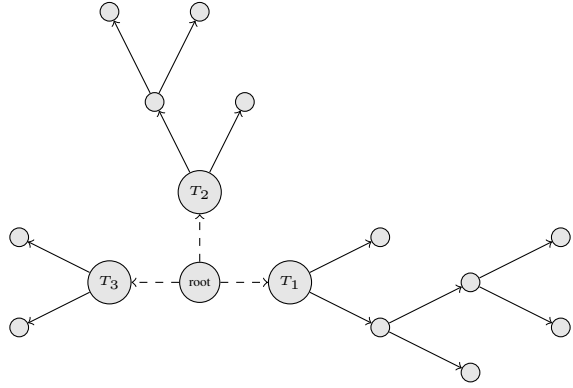

A forest is an ensemble of multiple decision trees $T_1,\ldots,T_K$.  
The forest shown in Figure~\ref{fig:forest} contains three trees $T_1$, $T_2$, and $T_3$. 
Each tree edge $e$ is associated with a variable $k_e$ and threshold $\thv_e$, and 
denotes a decision of the form $\Indnoarg(\bx[k_e] \leq \thv_e)$ or $\Indnoarg(\bx[k_e] > \thv_e)$.
Each node denotes a nonlinear decision rule of the form (\ref{eq:decision-rule}), which is the product 
of decisions along the edges leading from 
the root 
to this node.

Mathematically, each node $v$ of the forest is associated with a decision rule of the form
\[
 \basis_v(\bx) = \prod_{j}\Indnoarg(\bx[i_j] \leq \thv_{i_j})     \prod_{k}\Indnoarg(\bx[i_k] > \thv_{i_k})~, 
\]
which serves as a basis function or atom for the additive model considered in this paper.
Note that if $v_1$ and $v_2$ are the two children of $v$, then
$\basis_{v}(\bx)=\basis_{v_1}(\bx) + \basis_{v_2}(\bx)$. 
This means that any internal node is redundant 
in the sense that an additive model
with basis functions $\basis_v(\bx),\basis_{v_1}(\bx),\basis_{v_2}(\bx)$ can be represented as
an additive model over basis functions $\basis_{v_1}(\bx)$ and $\basis_{v_2}(\bx)$. 
Therefore it can be shown that an additive model over all tree nodes always has an 
{\em equivalent model} (equivalent in terms of output) over leaf nodes only. This property is important for computational efficiency because 
it implies that 
we only have
to consider additive models over leaf nodes. 

Let $\forest$ represent a forest, and each node $v$ of $\forest$ is associated with
$(\basis_v,\weight_v)$. 
Here $\basis_v$ is the basis function that this node represents;
$\weight_v$ is the {\em weight} 
or coefficient 
assigned to this node.  
The additive model of this forest $\forest$ considered in this paper is:
$
\func_{\forest}(\bx) = \sum_{v \in \forest} \weight_v \basis_v(\bx) 
$
with $\weight_v=0$ for any internal node $v$. 
In this setting, 
the regularized loss in 
(\ref{eq:reg-loss}) 
is a function of decision forest: 
\begin{equation}
\totalloss(\forest)= \cL(\func_{\forest}(X),Y) + \totalreg(\funcforest) . 
\label{eq:forest-loss}
\end{equation}

\subsection{Algorithmic framework}

The training objective of RGF is to build a forest that minimizes $\totalloss(\forest)$
defined in (\ref{eq:forest-loss}).  
Since the exact optimum solution is difficult to find,  we {\em greedily} select the {\em basis functions} and optimize the {\em weights}.
At a high level, we may summarize RGF in a generic algorithm in
Algorithm~\ref{alg:framework}. It essentially has two main components as follows.
\begin{itemize}
\item 
Fix the {\em weights}, and change the {\em structure} of the forest (which changes basis functions) so that the loss $\totalloss(\forest)$ is reduced the most 
(Line \ref{line:search}).  
\item
Fix the {\em structure} of the forest, and change the {\em weights} so that loss $\totalloss(\forest)$ is minimized
(Line \ref{line:optWeights}).  
\end{itemize}

%%--------------
\begin{algorithm}

\lnl{line:init}
    $\forest \assign \{\}$.  \\
    \Repeat{some exit criterion is met} {
\lnl{line:search}  
    $\forest \assign$
    the optimum forest that minimizes $\totalloss(\forest)$ 
    among all the forests that can be obtained by applying one step of structure-changing operation 
    to the current forest $\forest$. \\ 
\lnl{line:optWeights}  
      \lIf{some criterion is met}{ 
        optimize the leaf weights in $\forest$ to minimize loss $\totalloss(\forest)$.  
      }
    }
    Optimize the leaf weights in $\forest$ to minimize loss $\totalloss(\forest)$. \\
    \Return $\func_{\forest}(\mv{x})$ \\
\caption{Regularized greedy forest framework \label{alg:framework}}
\end{algorithm}
%%--------------

\subsection{Specific Implementation}
There may be more than one way to instantiate useful algorithms based on Algorithm~\ref{alg:framework}.
Below, we describe what we found effective and efficient.  

%---
\subsubsection{Search for the optimum structure change (Line \ref{line:search})} 
\label{sec:search}
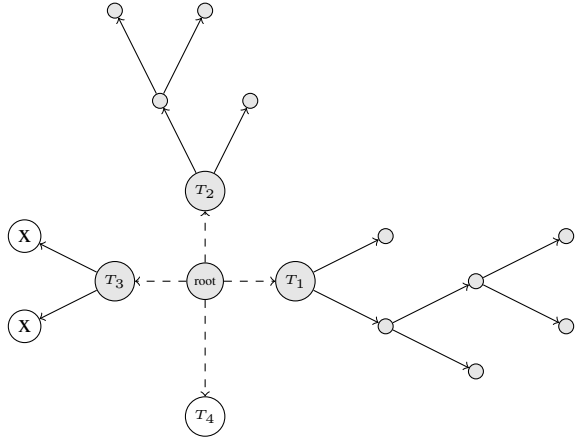
\begin{figure}[htb]
  \centering
  \begin{tikzpicture}
 [
    scale=1.2,
    observed/.style={circle,inner sep=2pt,draw=black,fill=black!10},
    front/.style={circle,inner sep=2pt,draw=black,fill=black!0}
    ]

   \node [observed,name=root] at (0,0) {\tiny root};

   \node[observed,name=T1] at (1,0){\tiny $T_1$};
   \draw[->,dashed] (root) to (T1);
   \node [observed,name=T11] at (2,0.5) {};
   \draw[->] (T1) to (T11);
   \node [observed,name=T12] at (2,-0.5) {};
   \draw[->] (T1) to (T12);
  \node [observed,name=T121] at (3,0) {};
   \draw[->] (T12) to (T121);
   \node [observed,name=T122] at (3,-1) {};
   \draw[->] (T12) to (T122);
   \node [observed,name=T1211] at (4,0.5) {};
   \draw[->] (T121) to (T1211);
   \node [observed,name=T1212] at (4,-0.5) {};
   \draw[->] (T121) to (T1212);

   \node[observed,name=T2] at (0,1){\tiny $T_2$};
   \draw[->,dashed] (root) to (T2);
   \node [observed,name=T21] at (0.5,2) {};
   \draw[->] (T2) to (T21);
   \node [observed,name=T22] at (-0.5,2) {};
   \draw[->] (T2) to (T22);
   \node [observed,name=T221] at (0,3) {};
   \draw[->] (T22) to (T221);
   \node [observed,name=T222] at (-1,3) {};
   \draw[->] (T22) to (T222);

   \node[observed,name=T3] at (-1,0){\tiny $T_3$};
   \draw[->,dashed] (root) to (T3);
   \node [front,name=T31] at (-2,0.5) {x};
   \draw[->] (T3) to (T31);
   \node [front,name=T32] at (-2,-0.5) {x};
   \draw[->] (T3) to (T32);

   \node[front,name=T4] at (0,-1.5){\tiny $T_4$};
   \draw[->,dashed] (root) to (T4) ;

  \end{tikzpicture}
%%  \vspace{-10pt}
  \caption{Decision Forest Splitting Strategy (we may either split a leaf in $T_3$ or start a new tree $T_4$) }
 \label{fig:forest-split}
\end{figure}

For computational efficiency, we only
allow the following two types of operations in the search strategy: 
\begin{itemize}
  \item to split an existing leaf node, 
  \item to start a new tree (i.e., add a new stump to the forest).  
\end{itemize}
The operations include assigning weights to new leaf nodes and setting zero to the node that was split. 
Search is done with the weights of all the existing leaf nodes fixed, by repeatedly evaluating the maximum loss reduction of all the possible structure changes.  
When it is prohibitively expensive to search the entire forest (and that is often the case with practical applications), we limit the search to the most recently-created $t$ trees with the default choice of $t=1$.  
This is the strategy in our current implementation. For example, 
Figure~\ref{fig:forest-split} shows that at the same stage as Figure~\ref{fig:forest}, we may either 
consider splitting one of the leaf nodes marked with symbol
$X$ or grow a new tree $T_4$ (split $T_4$'s root).

Note that RGF does not require the tree size parameter needed in GBDT.  With RGF, the size of each tree is automatically determined as a result of minimizing the regularized loss.

\newcommand{\aweight}{\weight}
\newcommand{\abasis}{\basis}
\newcommand{\one}{u_1}
\newcommand{\two}{u_2}
\newcommand{\onetwo}{u_k}
\newcommand{\basisone}{\basis_{\one}}
\newcommand{\basistwo}{\basis_{\two}}
\newcommand{\basisk}{\basis_{\onetwo}}

\newcommand{\dOne}[1]{#1'}   % first derivative
\newcommand{\dTwo}[1]{#1''}   % second derivative

\paragraph{Computation}
Consider the evaluation of loss reduction 
by splitting a node associated with $(\abasis,\aweight)$ into the nodes associated with $(\basisone,\aweight+\delta_1)$ and $(\basistwo,\aweight+\delta_2)$, 
and let us write $\newforest(\delta_1, \delta_2)$ for the new tree.  
Then the model associated with the new forest 
$\newforest(\delta_1,\delta_2)$
can be written as: 
\begin{equation}
  \func_{\newforest(\delta_1,\delta_2)}(\bx) 
   = \func_\forest(\bx) - \aweight \cdot \abasis(\bx)
   + \sum_{k=1}^2(\aweight+\delta_k)\basisk(\bx)
   = \func_\forest(\bx) + \sum_{k=1}^2 \delta_k \cdot \basisk(\bx) .
\label{eqn:newfunc}
\end{equation}
Recall that our additive models are over leaf nodes only.  
The node that was split 
is no longer leaf and therefore $\aweight \cdot \abasis(\bx)$ is removed from the model.  The second equality is from $\abasis(\bx)=\basisone(\bx)+\basistwo(\bx)$ due to the parent-child relationship.  
Note that, for the purpose of finding the optimum forest, we let 
$\newforest(\delta_1,\delta_2)$ go through all the possible forests 
that can be obtained by splitting one leaf node of the current forest $\forest$.  
However, 
our immediate goal here is to find
$\arg\min_{\delta_1,\delta_2} \totalloss(\newforest(\delta_1,\delta_2))$.                    
       
Actual computation depends on 
the loss function and the regularization term.  
In general, there may not be an analytic solution for this optimization problem, whereas 
we need to find the solution in an inexpensive manner as this computation is repeated frequently.  For fast computation, 
one may employ gradient-descent approximation as used in gradient boosting.
However, the sub-problem we are looking at is simpler, and thus instead of the simpler gradient descent approximation,
we perform one Newton step which is more accurate; namely, 
we obtain the approximately optimum $\hat{\delta}_k$ ($k=1,2$) as: 
\[
\hat{\delta}_k = 
- \frac{\dOne{\totalloss}_{\delta_k}(\newforest(\delta_1,\delta_2))}
       {\dTwo{\totalloss}_{\delta_k}(\newforest(\delta_1,\delta_2))}
       |_{\delta_1=0,\delta_2=0}~, 
\]
where $\dOne{\totalloss}_{\delta_k}(\cdot)$ and $\dTwo{\totalloss}_{\delta_k}(\cdot)$ are the first and second partial derivatives of 
$\totalloss(\cdot)$ with respect to $\delta_k$ ($k=1,2$).  
For example, with square loss and $L_2$ regularization penalty, i.e., 
$\totalloss(\forest)=\sum_{i=1}^n (h_\forest(\bx_i)-y_i)^2 + \lambda \sum_{v \in \forest} \weight_v^2$ 
with a constant $\lambda$, 
we have 
\[
  \hat{\delta}_k = 
\frac{
  \sum_{\basisk(\bx_i)=1}(y_i - \func_\forest(\bx_i)) - n\lambda\aweight
}
{
  \sum_{\basisk(\bx_i)=1}1 + n\lambda 
}
~, 
\]
which is the exact optimum for the given split.  

%%----
\subsubsection{Weight optimization/correction (Line \ref{line:optWeights})}
With the basis functions fixed, the weights can be optimized using a standard procedure if the regularization penalty is standard (e.g., $L_1$- or $L_2$-penalty).  
In our implementation we perform coordinate descent, which iteratively goes through the basis functions and in each iteration updates the weights by a Newton step with a small step size:
\begin{equation}
\weight_v \assign \weight_v - \eta \cdot 
\frac{ \dOne{\totalloss}_{\delta_v}(\forest(\delta_v)) }
     { \dTwo{\totalloss}_{\delta_v}(\forest(\delta_v)) }
     |_{\delta_v=0}
 , 
\label{eqn:optupdate}
\end{equation}
where $\delta_v$ is the additive change to $\weight_v$.  

Since the initial weights of new leaf nodes set in Line \ref{line:search} are approximately optimal at the moment, it is not necessary to perform weight correction in every iteration, which is relatively expensive.  
Based on the preliminary experiments using synthesized data, 
we found that correcting the weights every time $k$ new leaf nodes are added works well.   
%--
Obviously, $k$'s setting (the interval between fully-corrective updates) 
should not be extreme -- 
if $k$ is extremely large, it would be equivalent to doing fully-corrective update just once in the end 
and would lose the benefit of the interleaving approach; 
if $k$ is extremely small (e.g., $k=1$), it would slow down training. 
Empirically, as long as $k$ is not an extreme value, the choice of $k$ is not crucial.  
Therefore, 
we simply fixed $k$ to 100 in all of our experiments including the competitions we won, described later.  

%------
\subsection{Tree-structured regularization} 
\label{sec:reg}

Explicit regularization is a crucial component of this framework.  
To simplify notation,  
we define regularizers over a single tree.  
The regularizer over a forest can be obtained by adding the regularizers described here over all the trees.  
Therefore, 
suppose that we are given a tree $\tree$ with an additive model over leaf nodes: 
\[
  \func_{\tree}(\mv{x}) = \sum_{v \in \tree} \weight_v \basis_v(\bx)
  ~~,~~ \lw_v=0 \mbox{ for } v \notin \tleafset
\]
where $\tleafset$ denotes the set of leaf nodes in $\tree$. 

%%++++++++++++
\begin{figure}
\begin{center}
$\begin{array}{c}
\includegraphics[width=3in]{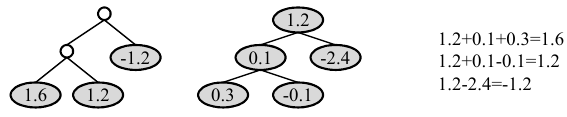}
\\
\end{array}$
\end{center}
%%\vspace{-20pt}
\caption{ \label{fig:equivalent} 
Example of Equivalent Models
}
\end{figure}
%%++++++++++++

To consider useful regularizers,   
first recall 
that for any additive model over leaf nodes only, there always exist {\em equivalent models} over all the nodes of the same tree that produce the same output. 
More precisely, let $\ancset{v}$ denote the set of ancestor nodes of $v$ and $v$ itself, 
and let $\tree(\anyww)$ be a tree that has the same topological structure as $\tree$ but whose node weights $\{\weight_v\}$ are replaced by $\{\anyww_v\}$.  
Then we have
\[
\forall u \in \tleafset: \sum_{\vx \in \ancset{u}} \anyww_\vx = \lw_u 
~~\Leftrightarrow~~
\func_{\tree(\anyww)}(\bx) \equiv \func_{\tree}(\bx)
\]
as illustrated in Figure \ref{fig:equivalent}.  
Our basic idea is that 
it is natural to give the same regularization penalty to all equivalent 
models defined on the same tree topology.  
One way to define a regularizer that satisfies this condition is to 
choose a model of some desirable properties as the unique representation for all the equivalent models and define the regularization penalty based on this unique representation.  
This is the high-level strategy we take.  
That is, 
we consider the following form of regularization: 
\[
\totalreg(\functree) = \sum_{v \in \tree(\regww)} \reg(v)
~:~~
\func_{\tree(\regww)}(\bx) \equiv \func_{\tree}(\bx)~. 
\]
Here node $v$ includes both internal and leaf nodes; 
the additive model 
$\func_{\tree(\regww)}(\bx)$
serves as the unique representation of the set of equivalent models; and 
$\reg(v)$ is a penalty function of $v$'s weight $\regww_v$ and $v$'s attributes such as the node depth.  
Each $\regww_v$ is a function of given leaf weights 
$\{ \lw_u \}_{u \in \tleafset}$, though the function may not be a closed form.  
Since regularizers in this form utilize the entire tree including its topological structure, we call them {\em tree-structured regularizers}.  
Below, we describe three tree-structured regularizers using three distinct unique representations.  

%%-----
\subsubsection{$L_2$ regularization on leaf-only models} 
\label{sec:rgfLeaf}

The first regularizer we introduce simply chooses the given leaf-only model as the unique representation 
and 
uses
the standard $L_2$ regularization.  This leads to 
a regularization term: 
\[
\totalreg(\functree)=\lambda \sum_{v \in \tree} \weight_v^2/2 
                = 
                \lambda \sum_{v \in \tleafset} \lw_v^2/2
\]
where $\lambda$ is a constant for controlling the strength of regularization.  
A desirable property of this unique representation is that 
among the equivalent models, the leaf-only model is often (but not always\footnote{
  For example, consider a leaf-only model on a stump whose two sibling leaf nodes have the same weight $\alpha \ne 0$.  Its equivalent model with the fewest basis functions (with nonzero coefficients) is the one whose weight is $\alpha$ on the root and zero on the two leaf nodes. 
}) the one with the smallest number of basis functions, 
i.e., the most sparse.  

%-----
\subsubsection{Minimum-penalty regularization}
\label{sec:treereg}

Another approach we consider is to choose the model that minimizes some penalty as the unique representative of all the equivalent models, as it is the most preferable model according to the defined penalty.  
We call this type of regularizer a {\em \regMiner}.  
In the following \regMiner, 
the complexity of a basis function is explicitly regularized via the node depth. 
\begin{equation}
\totalreg(\functree) = 
\lambda \cdot \min_{\{\ww_v\}} \left\{
  \sum_{\vx \in \tree} \frac{1}{2} \lamv{\vx} \ww_\vx^2:~
  \func_{\tree(\ww)}(\bx) \equiv \func_{\tree}(\bx)
\right\}~.  
\label{eqn:treereg}
\end{equation}
Here $d_\vx$ is the depth of node $\vx$, which is the distance from the root, 
and $\bdr$ is a constant.  A larger $\bdr>1$ penalizes deeper nodes more severely, 
which are associated with more complex decision rules, 
and we assume that $\bdr \ge 1$. 

%---
\paragraph{Computation}
To derive an algorithm for computing this regularizer, first 
we introduce auxiliary variables 
$\{\mww_\vx\}_{\vx \in \tree}$, recursively defined as: 
\[
    \mww_{\rt} = \ww_{\rt}~,~~
    \mww_\vx = \ww_\vx + \mww_{\parent{\vx}}~,
\]
where $\rt$ is $\tree$'s root, and $\parent{\vx}$ is $\vx$'s parent node, 
so that we have  
\begin{equation}
\func_{\tree(\ww)} \equiv \func_{\tree} \Leftrightarrow 
  \forall \vx \in \tleafset. \left[ \mww_\vx = \lw_\vx \right]~,
\label{eqn:mwwleafiff}
\end{equation} 
and (\ref{eqn:treereg}) can be rewritten as:  
\begin{align*}
\totalreg(\functree) &=
\lambda \cdot \min_{\{\mww_v\}} \left\{
   f(\{\mww_\vx\}): 
   \forall \vx \in \tleafset.[~\mww_\vx = \lw_\vx] 
\right\}  
\\
\mbox{ where }
f(\{\mww_\vx\}) &=
\sum_{\vx \ne \rt} \lamv{\vx}(\mww_\vx - \mww_{\parent{\vx}})^2/2
  + \mww_{\rt}^2/2~. 
\end{align*}
Setting $f$'s partial derivatives to zero, we obtain that 
at the optimum, 
\begin{equation}
\forall \vx \notin \tleafset:~ 
\mww_\vx = \left\{ \begin{array}{ll} 
  \frac{ \mww_{\parent{\vx}} + \sum_{\isChildOf{\wx}{\vx}} \bdr \mww_\wx }
       { \neighlam } 
  & \vx \ne \rt \\
  \frac{ \sum_{\isChildOf{\wx}{\vx}} \bdr \mww_\wx }
       { \neighlam }
  & \vx = \rt \\
\end{array} \right.~, 
\label{eqn:mwwopt}
\end{equation}
i.e., essentially, $\mww_\vx$ 
is the weighted average of the neighbors. 
This naturally leads to an iterative algorithm summarized in Algorithm \ref{alg:one}. 

%%--------------
\begin{algorithm}
\lFor{$\vx \in \tree$}{$
  \mww_{\vx,0} \assign \left\{ \begin{array}{ll} 
    \lw_\vx & \vx \in \tleafset \\
    0     & \vx \notin \tleafset 
  \end{array} \right.
$} 
\For{$i=1$ \KwTo $m$}{
  \lFor{$\vx \in \tleafset$}{$\mww_{\vx,i} \assign \lw_\vx$} 
  \lFor{$\vx \notin \tleafset$}{$
    \mww_{\vx,i} \assign \left\{ \begin{array}{ll}
    \frac{ \mww_{\parent{\vx},i-1} + \sum_{\isChildOf{\wx}{\vx}} \bdr \mww_{\wx,i-1} }
         { \neighlam } 
      & \vx \ne \rt \\
    \frac{ \sum_{\isChildOf{\wx}{\vx}} \bdr \mww_{\wx,i-1} }
         { \neighlam }
      & \vx = \rt \\
    \end{array} \right.
  $}
}
\Return $\{ \mww_{\vx,m} \}$ \\
\caption{
\label{alg:one}
}
\end{algorithm}

%%----------
\subsubsection{Min-penalty regularization with sum-to-zero sibling constraints}
\label{sec:treeregsib}

%%++++++++++++
\begin{figure}[h]
%\vspace{-15pt}
\begin{center}
$\begin{array}{c}
\includegraphics[width=1in]{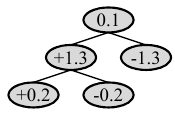}
\\
\end{array}$
\end{center}
%%\vspace{-20pt}
\caption{ \label{fig:sib} 
Example of Sum-to-zero Sibling Model
}
\end{figure}
%%++++++++++++

Another regularizer we introduce is based on the same basic idea as above but 
is computationally simpler.  
We add to (\ref{eqn:treereg}) the constraint that 
the sum of weights for every sibling pair must be zero, 
\[
\totalreg(\functree) = 
\lambda \cdot \min_{\{ \ww_v \}} \left\{
  \sum_{\vx \in \tree} \lamv{\vx} \ww_\vx^2/2:~
  \func_{\tree(\ww)}(\bx) \equiv \func_{\tree}(\bx)
  ;~
  \forall \vx \notin \tleafset.\left[ \sum_{\isChildOf{\wx}{\vx}} \ww_\wx=0 \right]
\right\} , 
\]
as illustrated in Figure \ref{fig:sib}.  
The intuition behind this sum-to-zero sibling constraints is that 
less redundant models are preferable and that 
the models are the  
{\em least redundant} when branches at every internal node lead to completely opposite actions, namely, `adding $x$ to' versus `subtracting $x$ from' the output value.  

Using the auxiliary variables $\{\mww_v\}$ as defined above, it is straightforward to show that any set of equivalent models has exactly one model that satisfies the sum-to-zero sibling constraints.  
This model
can be obtained through the following recursive computation
on the auxiliary variables: 
\[
\mww_\vx = \left\{ \begin{array}{ll}
  \lw_\vx & \vx \in \tleafset \\
  \sum_{\isChildOf{\wx}{\vx}} \mww_\wx/2 & \vx \notin \tleafset \\
\end{array} \right.
~. 
\]

%-------------
\subsection{Extension of regularized greedy forest}
We introduce an extension, which allows the process of forest growing and 
the process of weight correction to have different regularization parameters.   
The motivation is that 
the regularization parameter optimum for weight correction may not necessarily be optimal 
for forest growing, as the former is fully-corrective and therefore global whereas the 
latter is greedy and is localized to the leaf nodes of interest.  
Therefore, it is sensible to allow distinct regularization parameters for these two 
distinct processes.  
Furthermore, there could be an extension that allows one to change the strength of regularization 
as the forest grows, though we did not pursue this direction in the current work.

%*********
\newcommand{\tsize}{q} % for synthesized data generation

\newcommand{\nleaf}{\ell}
\newcommand{\nfeat}{d}
\newcommand{\ndata}{n}
\newcommand{\treesize}{z}
\newcommand{\optfrequency}{c}
%%------------------------------

%%-----
\section{Experiments}

This section reports empirical studies of RGF in comparison with GBDT 
and several tree ensemble methods.  
In particular, we report the results of entering competitions using RGF.  
Our implementation of RGF used for the experiments is 
available from \url{http://riejohnson.com/rgf\_download.html}.

%---
For clarity, 
the experiments focus on regression tasks and binary classification tasks.  
However, 
note that since the method is designed for optimizing general loss, there are 
other applicable tasks.  
For example,
multi-class categorization can be performed by combining binary classification tasks in the ``one-vs-others''
or other encoding schemes, as is commonly done with the methods that optimize general loss.  
In addition, there are multi-class training methods for, for example, GBDT and AdaBoost, and RGF can be extended similarly.  
%---

%%----
\subsection{On the synthesized datasets controlling complexity of target functions}
\label{sec:exp:syn}

%+++++++++++++++++++++++++++++++++++++++++++++++
\begin{table}\mytblsz
\begin{center}
\begin{tabular}{|c|c|c||cc|cc|} 
\hline
             & \rgfLeaf   & GBDT   & \multicolumn{4}{|c|}{RGF min-penalty}\\
             \cline{4-7}
             &            &        & \multicolumn{2}{|c|}{w/sib. constraint} & \multicolumn{2}{|c|}{w/o sib. constraint} \\
\hline
5-leaf data  & \bf{0.2200}& 0.2597 & 0.1885 & (0.0315) & 0.1890 & (0.0310) \\
10-leaf data & \bf{0.3480}& 0.3968 & 0.3270 & (0.0210) & 0.3266 & (0.0214) \\
20-leaf data & \bf{0.4578}& 0.4942 & 0.4545 & (0.0033) & 0.4538 & (0.0040) \\
\hline
\end{tabular}
\caption{
\label{tab:rsyn}
Regression results on synthesized datasets.  RMSE.  
Average of 3 runs, each of which used randomly-drawn 2K training data points.  
\rgfLeaf\ outperforms GBDT.  RGF min-penalty (with or without the sibling constraint)
further improves accuracy; the numbers in parentheses are 
accuracy improvements over \rgfLeaf.  
}
\end{center}
\end{table}
%+++++++++++++++++++++++++++++++++++++++++++++++

%%++++++++++++
\begin{figure}[h]
\begin{center}
\includegraphics[width=2in]{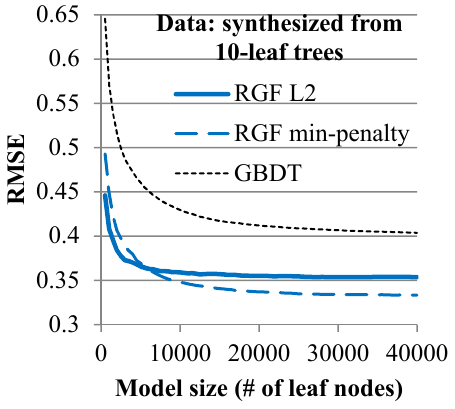}
\end{center}
\caption{ \label{fig:rsynfig}
Regression results in relation to model size.
One particular run on the data synthesized from 10-leaf trees.   
}
\end{figure}
%%++++++++++++

First 
we study the performance of the methods in relation to the complexity of target functions 
using synthesized datasets.  
To synthesize datasets, 
first we defined the target function by randomly generating 100 $\tsize$-leaf regression trees; then we randomly generated data points and applied the target function to them to assign the output/target values.  
%---
In more detail, 
(1) generate 100 trees of $\tsize$ leaf nodes by randomly choosing a node to split and also randomly choosing
features and threshold values for split; 
(2) assign weights $0, 1, \ldots, \tsize$ to the leaf nodes of each tree; 
(3) generate data points of 10 dimensions so that the components distribute uniformly over $\{0,1,\ldots,99\}$; 
(4) apply the tree ensemble generated above to each data point. The obtained value is an interim target
value.  To generate regression problems, normalize the interim target value by subtracting the mean and dividing
by the standard deviation.
%---
Note that a larger tree size $\tsize$ makes the target function more complex.  

The results shown in Table \ref{tab:rsyn} are in the root mean square error (RMSE) 
averaged over three runs.  In each run, randomly chosen 2K data points were used for training 
and the number of test data points was 20K.  
The parameters were chosen by 2-fold cross validation on the training data.  
Since the task is regression, the loss function for RGF and GBDT were set to square loss.  
RGF used here is the most basic version, which does $L_2$ regularization with one parameter 
$\lambda$ for both forest growing and weight correction.  
$\lambda$ was chosen from $\{1,0.1,0.01\}$.  
For GBDT, we used R package {\tt gbm}\footnote{
In the rest of the paper, {\tt gbm} was used for the GBDT experiments unless otherwise specified.  
}
\cite{Ridgeway:07a}.  
The tree size (in terms of the number of leaf nodes) and the shrinkage parameter were chosen 
from $\{5,10,15,20,25\}$ and $\{0.5,0.1,0.05,0.01,0.005,0.001\}$, respectively.  
Table \ref{tab:rsyn} shows that RMSE achieves smaller error than GBDT on all types of datasets.  

RGF with \regMiner\ with the sibling constraints 
further improves RMSE over \rgfLeaf\ by 0.0315, 0.0210, 0.0033 on the 5-leaf, 10-leaf, and 20-leaf 
synthesized datasets, respectively.  RGF with \regMiner\ without the sibling constraints also 
achieved the similar performances.  
Based on the amount of improvements, \regMiner\ appears to be more effective on simpler targets.  
Figure \ref{fig:rsynfig} plots RMSE in relation to the model size in terms of the number of basis 
functions or leaf nodes.
RGF produces better RMSE at all the model sizes; 
in other words, to achieve similar RMSE, RGF requires a smaller model than GBDT. 
  
The synthesized datasets used in this section are provided with the RGF software.  

%%-----
\subsection{Regression and 2-way classification tasks on the real-world datasets} 
\label{sec:exp:real}

%%++++++++++++
\begin{table}[h]\mytblsz
\begin{center}
\begin{tabular}{|c|c|l|}
\hline
Name         &Dim &\multicolumn{1}{|c|}{Regression tasks} \\
\hline
CT slices         & 384      & Target: relative location of CT slices \\
California Houses &   6      & Target: log(median house price) \\
YearPredictionMSD &  90      & Target: year when the song was released \\ 
\hline
\hline
Name         &Dim &\multicolumn{1}{|c|}{Binary classification tasks} \\
\hline
Adult        &  14(168) & Is income $>$ \$50K? \\
Letter       &  16      & A-M vs N-Z  \\
Musk         & 166      & Musk or not \\
Nursery      &   8(24)  & ``Special priority'' or not \\
Waveform     &  40      & Class2 vs. Class1\&3 \\
\hline
\end{tabular}
\end{center}
%%\vspace{-15pt}
\caption{ \label{tab:realworld_datasets} 
Real-world Datasets.  
We report the average of 3 runs, each of which uses 2K training data points. 
The numbers in parentheses indicate the dimensionality after converting categorical attributes to indicator vectors.  
}
\end{table}
%%++++++++++++

%+++++++++++++++++++++++++++++++++++++++++++++++
\begin{table}\mytblsz
\begin{center}
\begin{tabular}{|c|c|c|c|c|} 
\hline
         & \rgfLeaf & GBDT & RandomF. & BART \\
\hline
CT slices&\bf{7.2037}& 7.6861 &{\em 7.5029}&     8.6006 \\
California Houses   &\bf{0.3417}& 0.3454 &{\em 0.3453}&     0.3536 \\
YearPredictionMSD   &\bf{9.5523}& 9.6846 &     9.9779 &{\em 9.6126} \\
\hline
\end{tabular}
\caption{
\label{tab:rw-regress}
Regression results.  RMSE.  
Average of 3 runs, each of which used randomly-drawn 2K training data points. 
The best and second best results are in {\bf bold} and {\em italic}, respectively.  
}
\end{center}
\end{table}
%+++++++++++++++++++++++++++++++++++++++++++++++

%+++++++++++++++++++++++++++++++++++++++++++++++
\begin{table}\mytblsz
\begin{center}
\begin{tabular}{|c|ccc|ccc|c|c|ccc|} 
\hline
         & \multicolumn{3}{|c|}{\rgfLeaf}    & \multicolumn{3}{|c|}{GBDT}    & Random  & BART    & \multicolumn{3}{|c|}{AdaBoost}\\
         & Sq.       & Log.      & Expo.     & Sq.   & Log.      & Expo.     & forests &         & stumps & w/full & reg.\\
\hline
Adult    &     85.62 &     85.63 &     85.20 & 85.62 &{\bf 85.72}&{\em 85.75}& 85.29 &     85.62 &  85.51  &85.10 & 84.68\\
%Letter (40k)  &{\bf 92.59}&     92.39 &{\em 92.55}& 91.32 &     91.73 &     92.03 & 90.33 &     85.06 & 80.65  &79.67 & 92.12\\
Letter   &{\bf 92.50}&     92.19 &{\em 92.48}& 91.20 &     91.72 &     91.93 & 90.33 &     85.06 & 80.65  &79.67 & 92.12 \\
Musk     &     97.83 &{\bf 97.91}&{\em 97.83}& 97.14 &     96.79 &     97.27 & 96.23 &     95.56 & 96.91  &95.63 & 97.13 \\
Nursery  &     98.63 &{\bf 99.97}&{\em 99.95}& 98.13 &     99.90 &     99.85 & 97.44 &     99.12 & 93.31  &93.11 & 99.51 \\
Waveform &     90.28 &     90.21 &     90.06 & 89.56 &     89.97 &     90.18 & 90.20 &{\em 90.49}& 90.19  &88.08 &{\bf 90.52}\\
\hline
\end{tabular}
\caption{
\label{tab:rw-classif}
Binary classification results.  Accuracy (\%).  
Average of 3 runs, each of which used randomly-drawn 2K training data points. 
``Sq.'', ``Log.'', ``Expo.'' stand for the square loss, logistic loss, and exponential loss, respectively.  
``w/full'' is AdaBoost with stumps with fully-corrective update as post-processing.  
``reg.'' is AdaBoost with the regularized tree learner.  
The best and second best results are in {\bf bold} and {\em italic}, respectively.  
}
\end{center}
\end{table}
%+++++++++++++++++++++++++++++++++++++++++++++++

%%++++++++++++
\begin{figure}
\begin{center}
\includegraphics[width=4in]{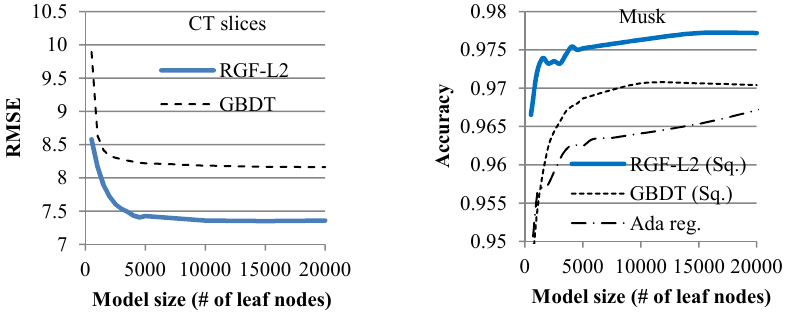}
\end{center}
\caption{ \label{fig:szfx} : 
RMSE/Accuracy in relation to model size.  
One particular run on the representative datasets.  
}
\end{figure}
%%++++++++++++

The first suite of real-world experiments use relatively small training data of 2K data points to facilitate experimenting 
with a wide variety of datasets.  
The criteria of data choice were (1) having over 5000 data points in total to ensure a decent amount 
of test data and (2) to cover a variety of domains. 
The datasets and tasks are summarized in Table \ref{tab:realworld_datasets}.  
All except Houses (downloaded from \url{http://lib.stat.cmu.edu}) are from the UCI repository \cite{Frank+Asuncion:2010}.  
All the results are the average of 3 runs, each of which used randomly-drawn 2K training data points.  
For multi-class data, binary tasks were generated as in Table \ref{tab:realworld_datasets}.  
The official test sets were used as test sets if any (Letter, Adult, and MSD).  
For relatively large Nursery and Houses, 5K data points were held out as test sets. 
For relatively small Musk and Waveform, in each run, 2K data points were randomly chosen as training sets, and the rest were used as test sets (4598 data points for Musk and 3000 for Waveform).  
The exact partitions of training and test data 
are provided with the RGF software.  
 
All the parameters were chosen by 2-fold cross validation on the training data.  
The RGF tested here is \rgfLeaf\ with the extension in which the processes of forest growing and weight 
correction can have regularization parameters of different values, 
which we call $\lambda_g$ (`$g$' for `growing') and $\lambda$, respectively.  
The value of $\lambda$ was chosen from $\{10,1,0.1,0.01\}$ with square loss, 
and from $\{10,1,0.1,0.01,1e-10,1e-20,1e-30\}$ with logistic loss and exponential loss.  
$\lambda_g$ was chosen from $\{\lambda, \frac{\lambda}{100}\}$.  
The tree size for GBDT was chosen from $\{5,10,15,20,25\}$, and the shrinkage parameter was from 
$\{0.5,0.1,0.05,0.01,0.005,0.001\}$.  

In addition to GBDT, we also tested two other tree ensemble methods: {\em random forests} \cite{Breiman+etal10}
and {\em Bayesian additive regression trees (BART)} \cite{ChGeMc10}.  
We used the R package {\tt randomForest} \cite{Breiman+etal10}
and performed random forest training with the number of randomly-drawn features in $\{ \frac{\nfeat}{4}, \frac{\nfeat}{3}, \frac{\nfeat}{2}, 
\frac{3\nfeat}{5}, \frac{7\nfeat}{10}, \frac{4\nfeat}{5}, \frac{9\nfeat}{10}, \sqrt{\nfeat} \}$, where $\nfeat$ is the feature dimensionality; 
the number of trees set to 1000; 
and other parameters set to default values.  
BART is a Bayesian approach to tree ensemble learning.  The motivation to test BART was that it 
shares some high-level strategies with RGF such as explicit regularization and non-black-box approaches to 
tree learners.  We used the R package {\tt BayesTree} \cite{ChiMc10} 
and chose the parameter $k$, which adjusts the degree of regularization, from $\{1,2,3\}$.  

Table \ref{tab:rw-regress} shows the regression results in RMSE.  
RGF achieves lower error than all others.  

Table \ref{tab:rw-classif} shows binary classification results in accuracy(\%).  
RGF achieves the 
best performance on the three datasets, whereas GBDT achieves the best performance on only one dataset.  

The \regMiner\ was found to be effective on Musk, improving the accuracy of \rgfLeaf\ with square loss 
from $97.83$\% to $98.39$\%, 
but it did not improve performance on other datasets.  Based on the synthesized data experiments 
in the previous section, we presume that this is because the target functions underlying 
these real-world datasets are mostly complex.  

%--  adaboost 
On the binary classification tasks, AdaBoost with three configurations was also tested\footnote{
Note that AdaBoost cannot be used for regression tasks since the loss function associated with AdaBoost is 
specifically the exponential loss. 
}: 
AdaBoost with decision stumps both with and without 
unregularized fully-corrective weight update to minimize exponential loss 
as post processing, 
and a publicly available AdaBoost implementation with tree ensembles.  
For the third configuration (labeled as `AdaBoost reg.' in the table) 
we used the R package {\tt ada} and set the parameter ``cp'', which controls the degree of regularization of the tree learner, from $\{0.1,0.01,0.001\}$ by cross validation.  
AdaBoost is a meta learner known to produce highly accurate classifiers, and in particular, 
AdaBoost with decision stumps has been intensively studied.  
The unregularized fully-corrective weight update of AdaBoost is discussed in the Appendix of \cite{ScSi99}.  

As shown in Table \ref{tab:rw-classif}, 
the accuracy of AdaBoost with decision stumps turned out to be generally poor, for example, 
the accuracy on Letter is about 12\% lower than the other methods.   
Among the three configurations of AdaBoost, `AdaBoost reg.' is the most competitive, which 
indicates that the success of the meta learner AdaBoost relies on the appropriate regularization of the base learner. 
Apparently, the degree of regularization implicitly provided by restricting the base learner to decision stumps 
is not the optimum on the three (Letter, Musk, and Nursery) out of five datasets, causing 
accuracy to degrade by 1\%, 6\%, and 12\% compared with RGF.  
The unregularized fully-corrective update (as suggested in \cite{ScSi99}) was found to degrade accuracy on all the datasets.  
This is not surprising because it is known that the exponential loss used in Adaboost is prone to overfitting, 
especially without regularization.
These AdaBoost results 
provide further support for our methodology of incorporating fully-corrective weight updates with explicit regularization.  

Regarding model sizes, 
we noticed that random forests and BART require far larger models than RGF to achieve 
the performances shown in the Table \ref{tab:rw-classif}; for example, all the BART's models consist 
of over 400K leaf nodes whereas the RGF models reach the best performance with 20K leaf nodes or fewer.  
Similarly, 
AdaBoost with stumps requires far larger models (200K leaf nodes) on Letter and Musk and yet 
it achieves lower accuracy than RGF.  
Fig. \ref{fig:szfx} shows the RMSE/accuracy of RGF and GBDT (and AdaBoost for classification) 
in relation to the model sizes on the representative datasets.  
Similar to Fig. \ref{fig:rsynfig} (on the synthesized data), RGF is more accurate than GBDT (and AdaBoost) 
at all model sizes; in other words, to achieve similar accuracy, RGF only requires a smaller model than GBDT.  

%%----
\subsection{GBDT with post processing of fully-corrective updates} 

%++++++++++++++++++++++
\begin{table}\mytblsz
\begin{center}
  \begin{tabular}{|c|r|rr|l|c|r|rr|}
  \multicolumn{5}{l}{Regression RMSE}   & \multicolumn{4}{l}{Classification accuracy}\\
  \cline{1-4} \cline{6-9}
                & GBDT & \multicolumn{2}{|c|}{GBDT w/post-proc.} & & & GBDT & \multicolumn{2}{|c|}{GBDT w/post-proc.} \\                              
  \cline{1-4} \cline{6-9}
  CT slices     &  7.6861     &\em{7.6371} & (2.9\%) & &   Adult  &     \em{85.62}&     85.10 & (13.5\%) \\
   Houses       & \em{0.3454} & 0.3753 & (11.8\%)    & &   Letter &     \em{91.20} &    90.69 & (8.7\%) \\ 
  MSD           & \em{9.6846} &10.0624 & (16.1\%)    & &   Musk   &     \em{97.14}&     96.62 & (7.3\%) \\
  \cline{1-4}
  \multicolumn{4}{l}{}                               & &   Nursery &    \em{98.13}&     97.62 & (8.8\%) \\
  \multicolumn{4}{l}{}                               & &   Waveform &   \em{89.56} &    88.71 & (12.8\%) \\
              \cline{6-9}
  \end{tabular}
\caption{
\label{tab:glmnet}
Comparison of GBDT with and without Fully-Corrective Post Processing proposed by \cite{FrPo08}; 
RMSE/accuracy(\%) and model sizes (in parentheses) relative to those without post-processing.    
Square loss.  Average of 3 runs.  
The post-processing decreases the model size, but noticeably degrades accuracy on all but one dataset.  
}
\end{center}
\end{table}
%++++++++++++++++++++++

A {\em two-stage} approach
was proposed in \cite{FriedmanPopescu03,FrPo08}\footnote{
  Although \cite{FrPo08} discusses various techniques regarding rules, 
  we focus on the aspect of the two-stage approach which \cite{FrPo08} 
  derives from \cite{FriedmanPopescu03}, since it is the most relevant portion 
  to our work due to its contrast with our interleaving approach.
}
that, in essence, first performs GBDT to learn basis functions and then fits their weights with $L_1$ penalty 
in the post-processing stage.  
Note that by contrast 
RGF generates basis functions and optimizes their weights in an {\em interleaving} manner so that fully-corrected weights can influence generation of the next basis functions.    

Table \ref{tab:glmnet} shows the performance results of the two-stage approach
on the regression and 2-way classification tasks described in Section
\ref{sec:exp:real}.  
As is well known, $L_1$ regularization has ``feature selection'' effects, assigning zero weights to more and more features with stronger regularization.  
After performing GBDT\footnote{
  We used our own implementation of GBDT for this purpose, as {\tt gbm} does not have the functionality 
to output the features generated by tree learning.  
}
with the parameter chosen by cross validation on the training data, 
we used the R package {\tt glmnet} \cite{FriHasTib11}
to compute the entire $L_1$ path in which the regularization parameter goes down gradually and thus more and more basis functions obtain nonzero weights, and chose the $L_1$ regularization parameter by 3-fold cross validation 
using the cross validation functionality of {\tt glmnet}.  
In the table, 
the numbers in the parentheses compare the sizes of the models with and without 
post-processing of the two-stage approach; for example, on Adult, the size of the model after post-processing is 13.5\% compared with the GBDT model without post-processing and accuracy is 0.52\% lower.  
The results show that the $L_1$ post processing makes the models smaller, 
but it noticeably degrades accuracy on all but one dataset.  
We view that for achieving better accuracy, RGF's interleaving approach has a clear advantage.  

%%----------
\subsection{RGF in the competitions}
\label{sec:competitions}

%+++++++++++++++++++++++++++++++++++++++++++++++
\begin{table}\mytblsz
\begin{center}
\begin{tabular}{|c|r|r|r|r|} 
\hline
             &\multicolumn{2}{|c|}{Data size}  & \multirow{2}{*}{dim}  & \multirow{2}{*}{\#team}\\
\cline{2-3} 
             &\#train  &\#test    &     & \\
\hline
Bond prices  & 762,678 &  61,146  &    91  & 265 \\
Bio response &   3,751 &   2,501  & 1,776  & 703\\
Health Prize &  71,435 &  70,942  & 50468  & 1,660\\
\hline
\end{tabular}
\caption{
\label{tab:competition}
Competition data statistics.  
The ``dim'' (feature dimensionality) and \#train are shown for the data used by one 
particular run for each competition for which we show the Leaderboard performance in Section \ref{sec:competitions}.  
}
\end{center}
\end{table}
%+++++++++++++++++++++++++++++++++++++++++++++++

To further test RGF in practical settings, we entered three machine learning competitions 
and obtained good results.  
The competitions were held in the ``Netflix Prize'' style. 
That is, participants submit predictions on the test data (whose labels are 
not disclosed) 
and receive performance results on the {\em public} portion of the test data as feedback
on the {\em public Leaderboard}.  
The goal is 
to maximize the performance on the {\em private} portion of the test data, and neither 
the private score nor the standing on the {\em private Leaderboard} is 
disclosed until the competition ends.  

In all of the three competitions, RGF produced more accurate models than GBDT.  
This demonstrates that RGF can achieve performance superior to GBDT even 
in the most competitive situation.  

%%++++++++++++
\begin{figure}
\begin{center}
\includegraphics[width=5in]{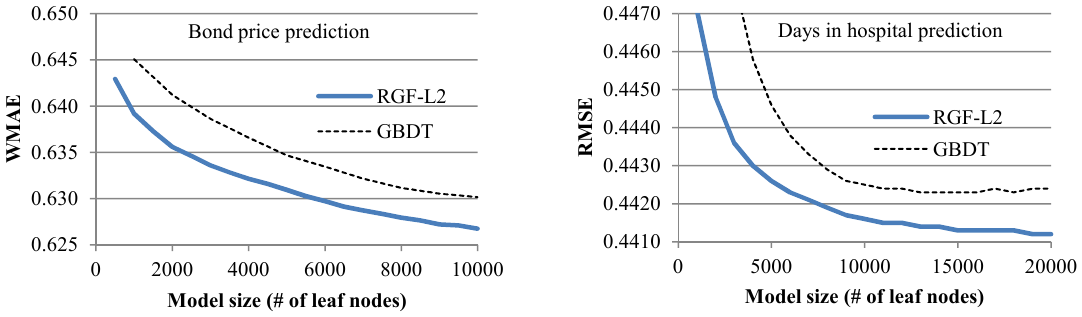}
\end{center}
\caption{ \label{fig:compe} : 
Accuracy in relation to model size.  
}
\end{figure}
%%++++++++++++

%%---
\subsubsection*{Bond price prediction}
We were awarded with the First Place Prize in Benchmark Bond Trade Price Challenge 
(\url{www.kaggle.com/c/benchmark-bond-trade-price-challenge}).  
The task was to predict bond trade prices based on the information 
such as past trade recordings.  

The evaluation metric was weighted mean absolute error, $\sum_i w_i|y_i - f(\mv{x}_i)|$, 
with the weights set to be larger for the bonds whose price prediction is considered to be harder.  
We trained RGF with L1-L2 hybrid loss \cite{BuLa97}, $\sqrt{1+r^2}-1$ where $r$ is the residual, 
which behaves like $L_2$ loss when $|r|$ is small and $L_1$ when $|r|$ is large.  
Our winning submission was the average of 62 RGF runs, 
each of which used different data pre-processing.

%%+++
\mytblszbegin
\begin{center}\begin{tabular}{|c|c|c|}
\hline
            & \multicolumn{2}{|c|}{Leaderboard WMAE} \\
\cline{2-3}
            & Public & Private \\
\hline
Our winning entry (Average of 62 \rgfLeaf\ runs) & \bf{0.68548}  & \bf{0.68031} \\
\hline
Second best team (GBDT and random forests) & 0.69380 & 0.69062 \\
\hline            
\hline
\rgfLeaf\ (single run) &\bf{0.69273}& \bf{0.68847} \\
\hline
GBDT (single run) & 0.69504 & 0.69582 \\
\hline
\end{tabular}\end{center}
\mytblszend
%%+++
  
In the table above, 
``\rgfLeaf\ (single run)'' is one of the RGF runs used to make the winning submission, and 
``GBDT (single run)'' is GBDT\footnote{
  As {\tt gbm} does not support the L1-L2 loss, 
  we used our own implementation.  
}
using exactly the same features as ``\rgfLeaf\ (single run)''.  
RGF produces smaller error than GBDT on both public and private portions.  
Furthermore, by comparison with the performance of the 
second best team (which blended random forest runs and GBDT runs), we observe that 
not only the average of the 62 RGF runs but also 
the single RGF run could have won the first place prize.
whereas the single GBDT run would have fallen behind the second best team.  

In Fig. \ref{fig:compe}, accuracy (in terms of WMAE) is shown in relation to model sizes 
on the 2-to-1 split of the data provided for training.  
RGF is more accurate than GBDT at all the model sizes; 
in other words, to achieve similar accuracy, RGF requires a smaller model than GBDT. 

%%---
\subsubsection*{Biological response prediction}
The task of Predicting a Biological Response 
(\url{www.kaggle.com/c/bioresponse})
was to predict a biological response (1/0) of molecules from their chemical properties.  
We were in the fourth place with a small difference from the first place.  

Our best submission combined the predictions of RGF and other methods with some data conversion.  
For the purpose of this paper, we show the performance of RGF and GBDT on the original data 
for easy reproduction of the results.  
Although the evaluation metric was log loss, 
$-\frac{1}{n} \sum_{i=1}^n y_i \log(f(\mv{x}_i)) + (1-y_i)\log(1-f(\mv{x}_i))$, 
we found that with both RGF and GBDT, better results can be 
obtained by training with square loss and then calibrating the predictions by: 
$g(x)=(0.05+x)/2 \mbox{ if }x<0.05; (0.95+x)/2 \mbox{ if }x>0.95; x \mbox{ otherwise }$.  
The log loss results shown in the table below were obtained this way.  
RGF produces better results than GBDT on both public and private sets.  
%%-
The same model size was used for both RGF and GBDT, which was found to generally produce the best accuracy for both.  
%%-
\mytblszbegin
\begin{center}\begin{tabular}{|c|c|c|}
\hline
     & \multicolumn{2}{|c|}{Leaderboard Log loss} \\
\cline{2-3}
     & Public & Private \\
\hline
\rgfLeaf & \bf{0.42224} & \bf{0.39361} \\
\hline
GBDT &     0.43576  &     0.40105  \\
\hline
\end{tabular}\end{center}
\mytblszend
%%---
\subsubsection*{Predicting days in a hospital -- \$3M Grand Prize}
After two kaggle competitions with good results, we decided to enter the 
highest profile kaggle competition at the time -- Heritage Provider Network Health Prize 
(\url{www.heritagehealthprize.com/c/hhp}).  
This was a two-year-long competition 
with \$3,000,000 Grand Prize and three milestones, 
which attracted 1660 teams.  
We entered the competition right before the 3rd/final milestone, in which we achieved the 2nd place, and then 
we were asked by other milestone winners to merge with them.  
The competition has concluded, and our team won the 1st place 
(though the threshold for the \$3M was not achieved). 

The task was to predict the number of days people will spend in a hospital in the next year based on 
``historical claims data''.  
We show the public Leaderboard performance of an RGF run and a GBDT run applied to 
the same features in the table below.  Both runs were part of the winning submission.
We also show the 5-fold cross validation results of our testing using training data on the same features.  
Again RGF achieves lower error than GBDT in both comparisons.

%%+++++
\mytblszbegin
\begin{center} \begin{tabular}{|c|c|c|}
\hline
     & Public LB RMSE & cross validation \\
\hline
\rgfLeaf & \bf{0.459877}  & \bf{0.440874} \\
\hline
GBDT &     0.460997   &     0.441612 \\
\hline
\end{tabular}\end{center} 
\mytblszend
%%+++++

Furthermore, we have 5-fold cross validation results of RGF and GBDT 
on 53 datasets each of which uses features composed differently.  
Their corresponding official runs are all part of the winning submission.
On {\em all} of the 53 datasets, RGF produced lower error than GBDT with the average of error differences 
0.0005, which is significant on this data.  
The superiority of RGF is
consistent on these datasets. This provides us 
competitive advantage to
do well in all three competitions we have entered.

In Fig. \ref{fig:compe}, accuracy (in terms of RMSE) is shown in relation to model sizes 
on the 4-to-1 split of the data provided for training.  
RGF is more accurate than GBDT at all the model sizes; 
in other words, to achieve similar accuracy, RGF requires a smaller model than GBDT. 

%%----------
\section{Running time}
\label{sec:time}

Compared with GBDT, computation of RGF involves additional complexity mainly for 
fully-corrective weight updates; however, running time of RGF is linear in the number of 
training data points.  
Below we analyze running time in terms of the following factors:
$\nleaf$, the number of leaf nodes generated during training; 
$\nfeat$, dimensionality of the original input space; 
$\ndata$, the number of training data points; 
$\optfrequency$, how many times the fully-corrective weight optimization is done; 
and 
$\treesize$, the number of leaf nodes in one tree, or tree size.  
In RGF, tree size depends on the characteristics of data and strength of regularization.  Although tree size can differ from tree to tree, for simplicity we treat it as one quantity, which should be approximated by the average tree size in applications.  

In typical tree ensemble learning implementation, for efficiency, 
the data points are sorted according to feature values at the beginning of 
training.  The following analysis assumes that this ``pre-sorting'' has been done.  
Pre-sorting runs in $O(\ndata\nfeat\log(\ndata))$, but its actual running time seems  
practically negligible compared with the other part of training even when $\ndata$ is 
as large as 100,000.  

Recall that RGF training consists of two major parts: 
one grows the forest, and the other optimizes/corrects the weights of leaf nodes.  
The part to grow the forest excluding regularization runs in $O(\ndata \nfeat \nleaf)$, same as GBDT. 
Weight optimization takes place $\optfrequency$ times, and each time
we have an optimization problem of $\ndata$ data points 
each of which has at most $\frac{\nleaf}{\treesize}$ nonzero entries; 
therefore, the running time for optimization, excluding regularization, is 
$O(\frac{\ndata \nleaf \optfrequency}{\treesize})$ using coordinate descent implemented with sparse matrix representation.   

%\paragraph{Regularization}
During forest building, the partial derivatives and the reduction of regularization penalty
are referred to $O(\ndata \nfeat \nleaf)$ times.  
During weight optimization, the partial derivatives of the penalty are required 
$O(\nleaf \optfrequency)$ times.  
With \rgfLeaf, 
computation of these quantities is practically negligible. 
Computation of \regMiners\ 
involves $O(\treesize)$ nodes; 
however, with efficient implementation that stores and reuses invariant quantities, 
extra running time for 
\regMiners\ 
during forest building can be reduced to 
$O(\ndata \nfeat \nleaf) + O(\nleaf \treesize^2)$ 
from  $O(\ndata \nfeat \nleaf \treesize)$.  
The extra running time during weight optimization is 
$O(\nleaf \optfrequency \treesize)$, but the constant part can be substantially reduced by efficient implementation.  

Actual execution time for training depends on not only the data but also 
the design of the parameter selection process.  
In our experiments in the previous section, 
we performed 2-fold cross validation for parameter selection from 8 (\rgfLeaf) and 30 (GBDT)  
parameter combinations for square loss; the total time 
for parameter selection on, for example, Letter, was 128 seconds with \rgfLeaf\ and 191 seconds with GBDT.  
That is, even though RGF training tends to take longer than GBDT individually, 
the total time for parameter selection could be shorter with RGF.  
On the same Letter dataset, parameter selection for AdaBoost with stumps (which was simply for deciding 
how large the model should be) took only 33 seconds; however, its accuracy is 12\% lower than RGF, which 
makes longer training time for RGF worthwhile.

%%-----------------
\section{Conclusion}

This paper introduced a new method that learns a nonlinear function by using an additive model over nonlinear decision rules. 
Unlike the traditional boosted decision tree approach, the proposed method directly works with
the underlying forest structure. The resulting method, which we refer to as regularized greedy forest (RGF), integrates two
ideas: one is to include tree-structured regularization into the learning formulation; and the other is to employ
the
fully-corrective regularized greedy algorithm.
Since in this approach we are able to take advantage of the special structure of the decision forest, 
the resulting learning method is 
effective and principled.  
Our empirical studies showed that 
the new method can achieve more accurate predictions 
than existing methods which we tested.

%\bibliographystyle{plain}
%\bibliography{greedy}
%--------------------------

%%-------------------

%%\vspace{-0.8in}
%%\begin{biography}[{\includegraphics[width=1in,height=1.25in,clip,keepaspectratio]{rjpic2}}]{Rie Johnson}
\begin{biography}[{\includegraphics[width=1in,height=1.25in,clip,keepaspectratio]{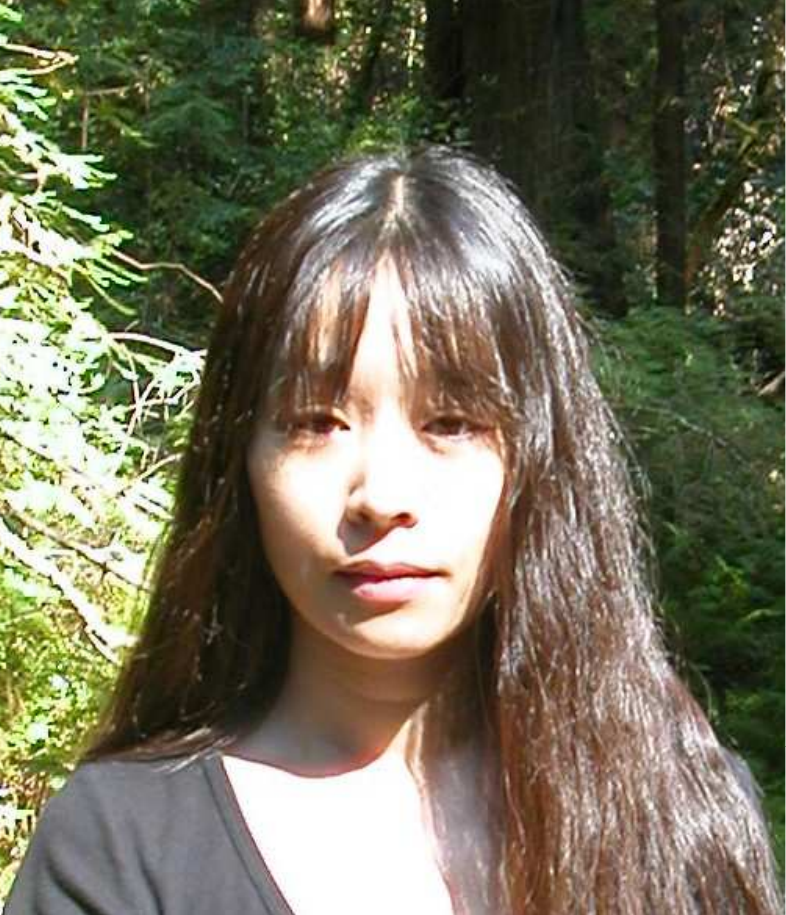}}]{Rie Johnson}
Rie Johnson obtained a PhD degree in computer science from Cornell University in 2001. She
was a research scientist at IBM T.J. Watson Research Center until 2007.
Her research interests are in machine learning and its applications.
\end{biography}
%%\vspace{-0.5in}
%%\begin{biography}[{\includegraphics[width=1in,height=1.25in,clip,keepaspectratio]{tzweb}}]{Tong Zhang}
\begin{biography}[{\includegraphics[width=1in,height=1.25in,clip,keepaspectratio]{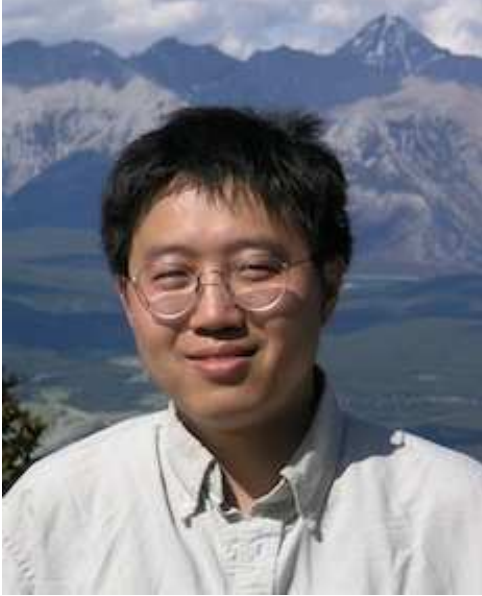}}]{Tong Zhang}
Tong Zhang received a B.A. in mathematics and computer
science from Cornell University in 1994 and a Ph.D. in Computer
Science from Stanford University in 1999. After graduation, he worked
at IBM T.J. Watson Research Center in Yorktown Heights, New York,
and Yahoo Research in New York city. He is currently a 
professor of statistics at Rutgers University. His research interests
include machine learning, statistical algorithms, their mathematical
analysis and applications.
\end{biography}

\end{document}